\documentclass[letterpaper, 10 pt, journal, twoside]{IEEEtran}  %

\IEEEoverridecommandlockouts                              

\usepackage{graphicx} 
\usepackage{amsmath} 
\usepackage{amssymb}  
\usepackage[usenames, dvipsnames]{color}

\usepackage{lipsum}
\usepackage{color}
\usepackage{cite}

\usepackage{diagbox}
\usepackage{tabu}
\usepackage{balance}  

\usepackage[ruled,linesnumbered]{algorithm2e}
\usepackage{algpseudocode}
\usepackage{epsfig}
\usepackage{multirow}

\usepackage{enumitem}
\usepackage{setspace}
\usepackage{tikz}
\usepackage{svg}
\usepackage{makecell}

\usepackage{fontenc}

\usepackage[acronym]{glossaries}
\setacronymstyle{long-short}
\newacronym{iou}{IoU}{intersection over union}
\newacronym{loocv}{LOOCV}{leave one out cross-validation}
\newacronym{ml}{ML}{machine learning}
\newacronym{dl}{DL}{deep learning}
\newacronym{rnn}{RNN}{recurrent neural network}
\newacronym{lstm}{LSTM}{long short-term memory}
\newacronym{convGRU}{convGRU}{convolutional gated recurrent unit}
\newacronym{gru}{GRU}{gated recurrent unit}
\newacronym{us}{US}{ultra sound}
\newacronym{fcn}{FCN}{fully connected network}
\newacronym{cnn}{CNN}{convolutional neural network}
\newacronym{tbtt}{TBTT}{truncated backpropagation through time}
\newacronym{bn}{BN}{batch normalization}
\newacronym{sgd}{SGD}{stochastic gradient descend}
\newacronym{bce}{BCE}{binary cross entropy}
\newacronym{relu}{ReLU}{rectified linear unit}

\usepackage{flushend}

\newcommand{\final}[1]{\textcolor{black}{#1}} 

\title{
DopUS-Net: Quality-Aware Robotic\\ Ultrasound Imaging based on Doppler Signal
}

\author{Zhongliang Jiang*, Felix Duelmer*, and Nassir Navab, \textit{Fellow, IEEE} 
\thanks{$^{*}$ Authors with equal contributions.}
\thanks{Z. Jiang, F. Duelmer, and N. Navab are with the Chair for Computer Aided Medical Procedures and Augmented Reality (CAMP), Technical University of Munich (TUM), 85748 Garching, Germany. {\tt\footnotesize{(zl.jiang@tum.de)}}
        }%
\thanks{This work involved human subjects in its research. Approval of all ethical and experimental procedures and protocols was granted by Institutional Review Board, No. 2022-87-S-KK, Declaration of Helsinki.}
}

\begin{document}

\maketitle


\begin{abstract}
Medical ultrasound (US) is widely used to evaluate and stage vascular diseases, in particular for the preliminary screening program, due to the advantage of being radiation-free. However, automatic segmentation of small tubular structures (e.g., the ulnar artery) from cross-sectional US images is still challenging. To address this challenge, this paper proposes the DopUS-Net and a vessel re-identification module that leverage the Doppler effect to enhance the final segmentation result. Firstly, the DopUS-Net combines the Doppler images with B-mode images to increase the segmentation accuracy and robustness of small blood vessels.
It incorporates two encoders to exploit the maximum potential of the Doppler signal and recurrent neural network modules to preserve sequential information. Input to the first encoder is a two-channel duplex image representing the combination of the grey-scale Doppler and B-mode \final{images} to ensure anatomical spatial correctness. The second encoder operates on the pure Doppler images to provide a region proposal. Secondly, benefiting from the Doppler signal, this work first introduces an online artery re-identification module to qualitatively evaluate the real-time segmentation results and automatically optimize the probe pose for enhanced Doppler images. This quality-aware module enables the closed-loop control of robotic screening to further improve the confidence and robustness of image segmentation. The experimental results demonstrate that the proposed approach with the re-identification process can significantly improve the accuracy and robustness of the segmentation results (dice score: from $0.54$ to $0.86$; intersection over union: from $0.47$ to $0.78$). 
The Code\footnote{Code: https://github.com/Felixduelmer/DopUs} and  Video\footnote{Video: https://www.youtube.com/watch?v=ZH7K63GngdA} are publicly accessible.
\end{abstract}

\begin{figure}[ht!]
\centering
\includegraphics[width=0.48\textwidth]{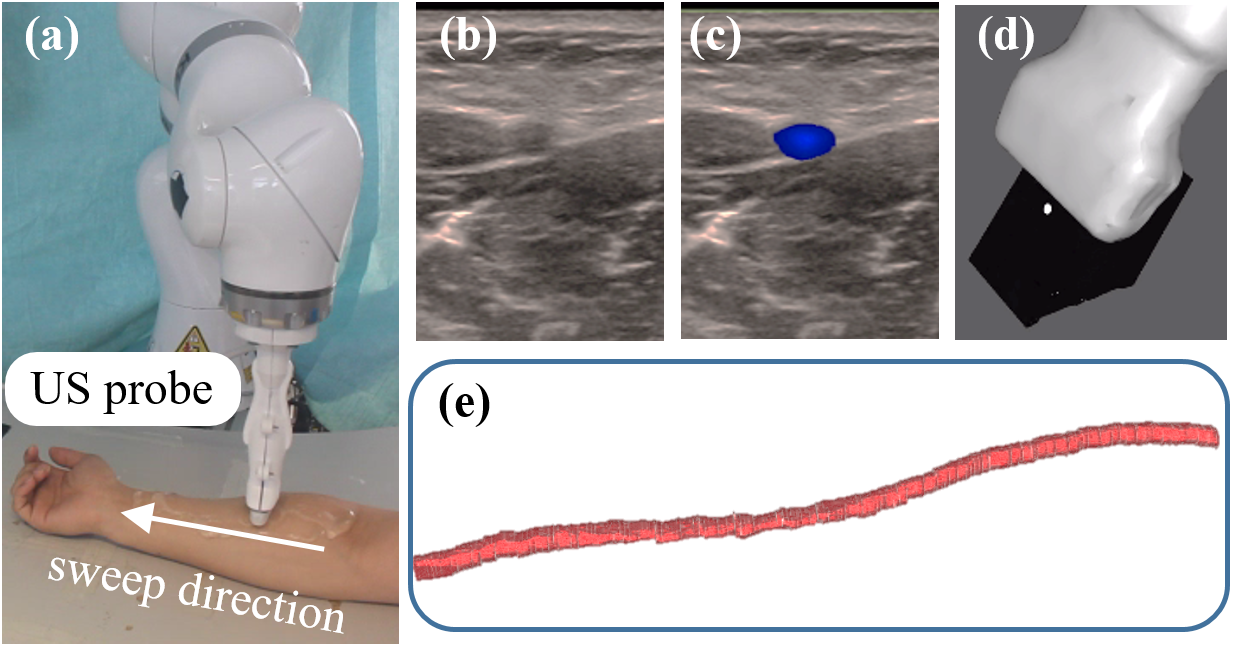}
\caption{(a) A scene of robotic US scanning of a volunteer's arm; (b) and (c) are B-mode and corresponding color Doppler images, respectively; (d) is the 3D view of tracked binary segmentation results in real-time; (e) is the reconstructed 3D vessel of interest.
}
\label{Fig_demo_question}
\end{figure}

\def\abstractname{Note to Practitioners}
\begin{abstract}
The Doppler signal is important for the diagnosis of vascular disease, e.g., peripheral arterial disease, in clinical practices, nevertheless it is not of similar significance for state-of-the-art robotic ultrasound (US) examination systems yet. This paper explores various neural network structures to effectively extract the blood vessels from US images by incorporating the Doppler signal into the segmentation process. The final DopUS structure with two encoders extracting differentiated information from two different inputs and fusing the latent feature representations in the bottleneck layer can also inspire other tasks like multi-senor fusion.
In addition, this work developed a Doppler-based tracker to assess the quality of the segmentation results in real-time. The assessment is subsequently used for a quality-aware module that enables closed-loop control of the robotic screening. Preliminary physical experiments suggest that the quality-aware robotic screening system can improve the confidence and robustness of autonomous US examination results. In the future, the Doppler signal could also be used to support clinical diagnosis. We believe the proposed quality-aware autonomous screening system is important for the development of large-scale robotic US screening programs. It will not only benefit the examination of limb arteries but also other vascular structures, e.g., carotid or aorta. 
\end{abstract}





\bstctlcite{IEEEexample:BSTcontrol}

\section{Introduction}
\IEEEPARstart{P}{eripheral} arterial disease (PAD) refers to the pathological process causing obstruction to blood flow in the arteries. It is a chronic disease of the aortic, iliac, and limb arteries (see Fig.~\ref{Fig_demo_question})~\cite{hankey2006medical}. Stenosis is one of the most common PAD, which narrows the blood vessel due to the plaque building up inside arteries; thereby, restricting the supply of blood. This can result in stroke and amputation in the worst cases.
PAD affects approximately $20\%$ of the population older than $55$ years~\cite{hirsch2001peripheral}.
However, the patient awareness of PAD diagnosis is low, particularly for the cases with mild symptoms, e.g., numbness, muscle shrinking, and ulcers~\cite{hirsch2001peripheral}. Thereby, a regular screening program for PAD will benefit early detection, which can further lead to the improvement of long-term life quality and a decrease in systemic cardiovascular risk.

\par
Preliminary screening for PAD is often performed in primary care practices using the ankle-brachial index (ABI). The ABI examines the blood pressure in the extremities but may result in an underdiagnosis of PAD~\cite{hirsch2001peripheral}. Although the ABI method is cheap, it cannot provide the location of the stenosis and is highly user-dependent. Conversely, computed tomography angiography (CTA), or magnetic resonance angiography (MRA)~\cite{rocha2014peripheral} can provide accurate clinical information. However, MRA is costly and CTA has ionizing radiation. Therefore, both of them are not suitable for extensive preliminary screening programs, where a large part of the participants are healthy. Considering the aforementioned constraints, ultrasound (US) imaging has been seen as a promising alternative for examining the extremity artery tree due to the absence of contraindications~\cite{favaretto2007analysis}. Furthermore, US has been extensively utilized for visualizing internal lesions and organ abnormalities due to its ability to provide rapid and straightforward diagnosis~\cite{guo2018ultrasound}. Regarding PAD, US imaging can provide information on the degree of calcification~\cite{favaretto2007analysis}. Nevertheless, nonnegligible inter- and intra-operator variations limit the clinical acceptance of the traditional US examination. 

\par
To tackle this shortage, robotic US systems (RUSS) have been seen as a promising solution to provide accurate, stable, and clinician-independent diagnosis results~\cite{gilbertson2015force, pierrot1999hippocrate, jiang2020automaticTIE, tan2022flexible, jiang2020automatic}. Specific to the RUSS developed for screening tubular structures, Janvier~\emph{et al.} employed a 6-DoF industrial robotic arm to hold a US probe and evaluated the performance of their proposed RUSS on a lower-limb mimicking phantom~\cite{janvier2008performance}. The experimental results demonstrated that the RUSS could be of value for the clinical evaluation of lower limb vessels over long and tortuous segments below the knee by accurately identifying the stenosis section. Compared to the traditional 2D images, 3D images are more intuitive for clinicians to identify lesion locations based on the reconstructed 3D anatomy.
To characterize the geometry of the vessel, Merouche~\emph{et al.} segmented vessel lumen using a fast marching method based on gradients~\cite{merouche2015robotic}. However, the efficiency of this approach is limited as the process takes $15$ minutes to scan $156~mm$ of a femoral artery. Since the aforementioned work was validated using a phantom without any blood flow, no color Doppler images are present. Yet, Doppler imaging is widely used in real scenarios for vascular examination and provides valuable information to facilitate both blood vessel detection and clinical analytical diagnosis. 

\par
To bridge this gap, we propose a novel RUSS combining Doppler with B-mode images to facilitate the real-time and accurate segmentation of small vessels like the radial artery (mean$\pm$SD diameter: $2.4\pm0.4$~mm~\cite{beniwal2014size}). To fully take advantage of the Doppler effect, a deep neural network (DopUS-Net) consisting of two encoders and one decoder is proposed. The Doppler image is first concatenated to the synchronized B-mode image to form a two-channel image to be used as one of the inputs to the DopUS-Net. The other input is the individual Doppler image, which is further used to provide coarse vessel locations and therefore acts as a region proposal module to increase the overall robustness and reliability. 
The main contributions of this work are summarized as follows:

\begin{itemize}
  \item We explore the optimal way to take advantage of the Doppler signal beside B-mode images to facilitate the accurate and robust segmentation of small limb arteries. The DopUS-Net with two encoders is proposed to effectively fuse the Doppler and B-mode images. The two encoders with two-channel duplex images (Doppler and B-mode) and pure Doppler images are designed to accurately extract the artery boundary and provide the coarse region proposal, respectively. 

  \item We first present an online segmentation quality-aware module based on the Doppler signal. Such a module enables close monitoring of imaging quality of robotic screening to improve the confidence and robustness of image segmentation, which can effectively improve the accuracy and completeness of the reconstructed 3D vessel. 
  
  \item We visualize the tubular structure of interest in a 3D view [see Fig.~\ref{Fig_demo_question}~(e)] to facilitate the intuitive assessment of clinicians. Due to the short segmentation time ($9~ms$), the 3D visualization process can be seen as real-time.
  

  

\end{itemize}
The experiments are performed on seven healthy volunteers and the experimental results demonstrate that the proposed DopUS-Net can outperform existing methods~\cite{ronneberger2015u,jiang2021automatic_baichuan} in terms of dice score. The online evaluation of the Doppler signals' stability and quality helps to improve the robustness and quality of the resulting 3D compounding image.

\par
The rest of this paper is organized as follows. Section II presents related work. The dataset preparation and the implementation details of the DopUS-Net are presented in Section III. Section IV describes the details of robotic scanning and the online artery re-identification approach. The experimental results on seven volunteers are provided in Section V. Finally, the summary of this study is presented in Section VI.

\section{Related Work}
\subsection{Feature-based Ultrasound Vessel Segmentation}~\label{sec:related_vessel}
\par
Compared to other popular medical imaging modalities, e.g., CT or MRI, US often suffers from inconsistent quality, speckle, and artifacts due to the intrinsic physics of wave propagation, like interference and scattering effects~\cite{mishra2018ultrasound}. To optimize the acoustic coupling performance, sonographers need to carefully adjust the pressure and probe orientation to avoid acoustic shadow and improve the imaging contrast. These factors contribute to making US one of the most challenging modalities for robust and accurate segmentation. Regarding the segmentation of blood vessels from cross-sectional US images, the Frangi filter~\cite{frangi1998multiscale} was developed based on the Hessian matrix in the early studies. Such an approach allows real-time segmentation, but presents limited segmentation accuracy.

\par
Since the shape of vascular structures on cross-sectional images are close to the ellipse, Smistad~\emph{et al.} proposed a template-based approach to automatically segment the object of interest~\cite{smistad2015real}. Yet, this approach is limited with regard to the segmentation accuracy because the segmented shapes are forced to be ellipse. To further enable accurate detection of vessel boundaries, Karami~\emph{et al.} proposed an adaptive polar active contours method to segment the jugular vein~\cite{karami2018adaptive}. They introduced a set of energy functions to consider local information, which enables robust segmentation of vessels even when the imaging quality is poor. However, this method requires a manual initialization of the object on the first frame. 
In addition, Abolmaesumi~\emph{et al.} compared the performance of five popular feature tracking approaches (the cross-correlation, the sequential similarity detection, the Star algorithm, the Star-Kalman algorithm, and the discrete snake algorithm) on carotid arteries~\cite{abolmaesumi2002image}. The experimental results demonstrated that the Similarity Detection method and the Star-Kalman algorithm can achieve noted tracking performance, while the Correlation and Star algorithms result in poorer performance with higher computational cost.

\par
\subsection{Learning-based Ultrasound Vessel Segmentation}
\par
Compared with traditional feature-based approaches, learning-based approaches have been seen as a promising alternative for the accurate segmentation of US images in recent years. Thanks to the inventions related to the convolutional neural network (CNN), learning-based approaches have achieved phenomenal success on various computer vision tasks as well as medical imaging segmentation~\cite{litjens2017survey}. Specific to biomedical image segmentation, U-Net and its variants have been widely used~\cite{ronneberger2015u, jiang2021autonomous, huang2021towards}. Such U-shape networks are based on fully convolutional neural networks (FCN). Multiple skip connections are used to pass the information from the encoder to the decoder assisting in accurate boundary extraction. To achieve automatic and accurate segmentation of US images, Mishra~\emph{et al.} proposed a FCN trained to learn structural boundary definitions to extract vessels from US images~\cite{mishra2018ultrasound}. Considering that the biomedical anatomy is continuous, Chen~\emph{et al.} employed a recurrent unit to preserve the historical information to assist the real-time segmentation~\cite{chen2020deep}. To improve generality, the salient image features were extracted from each spatial resolution level within the encoder-decoder structure. Yet, it is still challenging to accurately segment small vessels from B-mode images. To tackle this problem, B. Jiang~\emph{et al.} used both B-mode and color Doppler images to train the VesNetSCT++ for small vessels, e.g., femoral and tibial artery~\cite{jiang2021automatic_baichuan}. To effectively leverage the spatiotemporal context in the image sequences to improve the segmentation of small-scale arteries, VesNetSCT+ incorporates temporal, spatial, and feature-aware contextual embedding from two-channel images consisting of B-mode and Color Doppler images. Compared with classic approaches, learning-based approaches demonstrated advantages in time efficiency and segmentation accuracy.

\par
\subsection{Robotic US Screening System}
\par
The traditional free-hand US examination suffers from inter- and intra-operator variations, which significantly impair the clinical acceptance of the US modality. Benefiting from the controllable robotic mechanism, accurate and repeatable US images can be achieved by properly tuning the acquisition parameters. The characteristic of reproducibility is crucial for clinical applications requiring long-term care, e.g., monitoring the changes of vascular plaque or internal tumors~\cite{pheiffer2014model}. In addition, the development of RUSS has the potential to relieve sonographers from tedious and burdensome workloads, thereby reducing work-related musculoskeletal disorders.

To achieve reproducible US images, similar acquisition parameters (i.e., contact force, and probe orientation) are necessary across multiple US sweeps, yet, this is challenging even for experienced sonographers. The advantage of multiple sensing sources and precise adjustment of servomotors allows robotic manipulators to accurately repeat the US acquisitions. To obtain high-quality US images, Gilbertson~\emph{et al.} presented a compliant controller for a one degree of freedom (DoF) mechanism to stabilize images during examinations~\cite{gilbertson2015force}. Pierrot~\emph{et al.} proposed a hybrid force/position controller for a 7-DoF RUSS based on an external force/torque sensor~\cite{pierrot1999hippocrate}. A low-level embedded joint controller incorporating a PID algorithm was employed. The outer control loop, driven by the PID algorithm, was executed on an external workstation and utilized either force or position as the reference variable. This significantly improved the adaptability to other robotic manipulators. Regarding the optimization of probe orientation, Z. Jiang~\emph{et al.} quantitatively measured the effects of probe orientation on the resulting images and proposed a mechanical, model-based, approach to automatically identify the normal direction of an unknown constrained surface~\cite{jiang2020automaticTIE}. Due to the intrinsic physics of wave signal propagation, the orthogonal orientation of the probe can lead to US images with higher contrast because more signals can be reflected back to the US probe rather than scattering away~\cite{ihnatsenka2010ultrasound}. To provide stable image quality during scans, particularly for soft tissue like the breast, Tan~\emph{et al.} proposed a flexible RUSS and an online force adjustment approach based on real-time image feedback~\cite{tan2022flexible}.

\par
To quantify limb arterial stenosis, Janvier~\emph{et al.} used a 6-DoF robotic manipulator to control and standardize the 3D US acquisition process for large scanning distances~\cite{janvier2008performance}. The scanning path is generated in manual teaching mode for individual patients. To identify the stenosis location, the inner diameters of the vascular phantom were computed. Janvier~\emph{et al.} further proved clinical feasibility on volunteers by comparing 3D vascular volumes computed from B-mode and Doppler images, respectively, to the pre-scanned CTA~\cite{janvier20143}. To realize autonomous scanning, Merouche~\emph{et al.} moved the probe in a given direction step by step and computed the in-plane movement based on the segmented lumen to centralize the vessel of interest~\cite{merouche2015robotic}. Based on both in-vitro and in-vivo validations, the feasibility of applying RUSS to reconstruct limb vessels in a clinical context was investigated. 

\par
To further eliminate the requirements of manual selection of the scanning start and end point, Virga~\emph{et al.} employed a surface registration to transfer a generic scanning path planned on a preoperative MRI to the current environment for autonomous aortic screening~\cite{virga2016automatic}. To guarantee the overall imaging quality, they optimized the contact force by maximizing the overall confidence value of the resulting images. The US confidence value is often used to approximate the strength of US signal at each pixel, which can be computed based on~\cite{karamalis2012ultrasound}. Since the screening trajectory was fixed during the scanning, the resulting image quality decayed significantly if the objects moved during the scan. Regarding the examination of objects that extend over a large distance, i.e., limb arteries, sonographers even need to actively adjust patients' limbs to fully visualize the complete arterial tree. To tackle this practical challenge, Z. Jiang~\emph{et al.} proposed a motion-aware system based on an RGB-D camera and passive markers to monitor and compensate for potential movements of the patient during the scans~\cite{jiang2021motion}. 

\par
Recently, Z. Jiang~\emph{et al.} presented an end-to-end RUSS to automatically scan limb arteries based on the real-time image feedback~\cite{jiang2021autonomous}. The probe orientation adjustment was calculated, estimating the local vascular diameters and consequently solving a set of modeled optimization equations. The results demonstrated that their approach can effectively improve the accuracy and stability of scans compared with the free-hand US manner. Huang~\emph{et al.} imitated clinical protocols to automatically move the probe along the longitudinal direction of the carotid artery~\cite{huang2021towards}. 
To automatically search for the standard longitudinal plane of tubular structures, Bi~\emph{et al.} proposed a reinforcement learning network~\cite{bi2022vesnet}. The segmented binary masks generated by a classic U-Net~\cite{ronneberger2015u} were used as the state representation. This can bridge the gap between the simulated training environment and the real scenario, thereby achieving good generalization ability.







\section{Vessel Segmentation}
\par
Accurate segmentation of tubular structures from cross-sectional B-mode images is crucial for achieving accurate geometry of the vessel of interest; thereby, realizing accurate diagnosis and evaluation of PAD. Due to the difficulties implied by US artifacts such as speckle and occlusion, it is challenging to obtain a robust, reliable, and repeatable segmentation process~\cite{mishra2018ultrasound}. This limits the development of the autonomous screening program for PAD. To improve the segmentation performance, we propose a DopUS network using two encoders and one decoder to fully take advantage of the two modalities present in the duplex US images. Finally, the proposed DopUS Network is compared to the standard U-Net and other reference structures.


\subsection{Dataset}~\label{sec:dataset}
\subsubsection{US Data Recording}
\par
In this work, all US images were recorded from an ACUSON Juniper US machine (Siemens Healthineers, Germany) using a linear probe 12L3 (Siemens Healthineers, Germany, acquisition width: $51.3~mm$). To access the duplex images, a frame grabber (Epiphan Video, Canada) was used to connect the US machine and the main workstation via a USB interface. Due to the generation of color Doppler images, a small recording frequency ($10~fps$) was used. Since the main limb arteries are located close to the skin surface, the image depth for the transducer was set to $45~mm$. The other acquisition parameters were set as follows: Tissue Harmonic Imaging (THI): $8.4~MHz$, Dynamic Range (DynR): $75~dB$, US imaging focus: $20~mm$. The focus defines the region with the highest imaging quality on the resulting images. The THI is the method often used to increase contrast resolution, and the DynR represents the difference between the largest and smallest signals, which governs the images' gray scale levels. 

\par
The US images were recorded from seven volunteers including two females and five males. For sweep recording, the volunteers were asked to sit comfortably and put their left arm on a flat table. Every patient was scanned two times from the start of the brachial artery at the inner side of the shoulder joint towards the wrist. Both scans include the section from the shoulder to the bifurcation of the brachial artery into the radial and ulnar arteries. Thereafter, one scan followed the ulnar artery, while the other focused on the radial artery. Thus, the major arteries of the upper limb are present in the dataset. 

\par
The scans were carefully evaluated and manually annotated under the supervision of an experienced physician. As a consequence two of them were discarded due to improper image quality. A total of $12$ sweeps with $400-500$ images per sweep were used for the dataset. Due to the fact that a recurrent structure is used in the network, sequential data is required. The sweeps were therefore split into $279$ sequences with a fixed length of $20$ images/sequence. All in all $5580$ labeled images, with corresponding color Doppler images and B-mode images, were used for training.

\begin{figure}[ht!]
\centering
\includegraphics[width=0.45\textwidth]{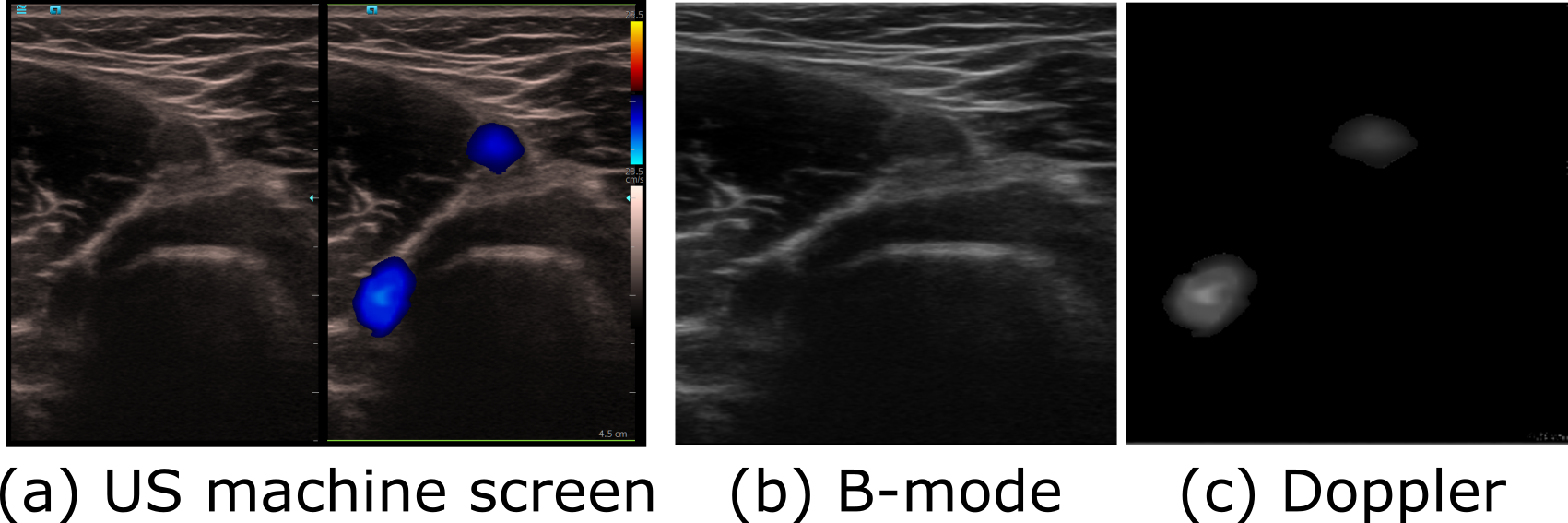}
\caption{Illustration of the pre-processing pipeline.
}
\label{fig:pre_processing_pipeline}
\end{figure}

\subsubsection{Duplex Data Pre-Processing}
\par
The frame grabber can only provide access to the same images displayed on the US machine screen \final{Fig.}~\ref{fig:pre_processing_pipeline}~(a). In order to segment the US images and further visualize the segmented binary masks in 3D in real-time, online pre-processing was necessary. The received images [Fig.~\ref{fig:pre_processing_pipeline}~(a)] were processed to generate B-mode images [Fig.~\ref{fig:pre_processing_pipeline}~(b)] and Doppler images [Fig.~\ref{fig:pre_processing_pipeline}~(c)], respectively. To extract Doppler features, the recorded images were converted to the HSV color space. Thereby, the colored Doppler features (blue or red in Fig.~\ref{fig:pre_processing_pipeline}~(a)) can be easily extracted by setting a threshold in terms of saturation ($\geq 100$) and value ($\geq 20$). To reduce the number of input parameters, both duplex images were compressed to $320 \times 320$ pixels for the segmentation network. The original size of the images was $497 \times 733$ pixels. It should be noted that this downsampling procedure may introduce some degree of distortion and possible information loss. However, we deem it acceptable given that the blood vessels in question occupy a contiguous area that remains discernible in the lower-resolution representation. The implementation of the pre-processing pipeline was done using the open-source library OpenCV~\cite{bradskiOpenCVLibrary2000}. 



\subsection{Network Architecture}
\par

\begin{figure*}[ht!]
\centering
\includegraphics[width=0.85\textwidth]{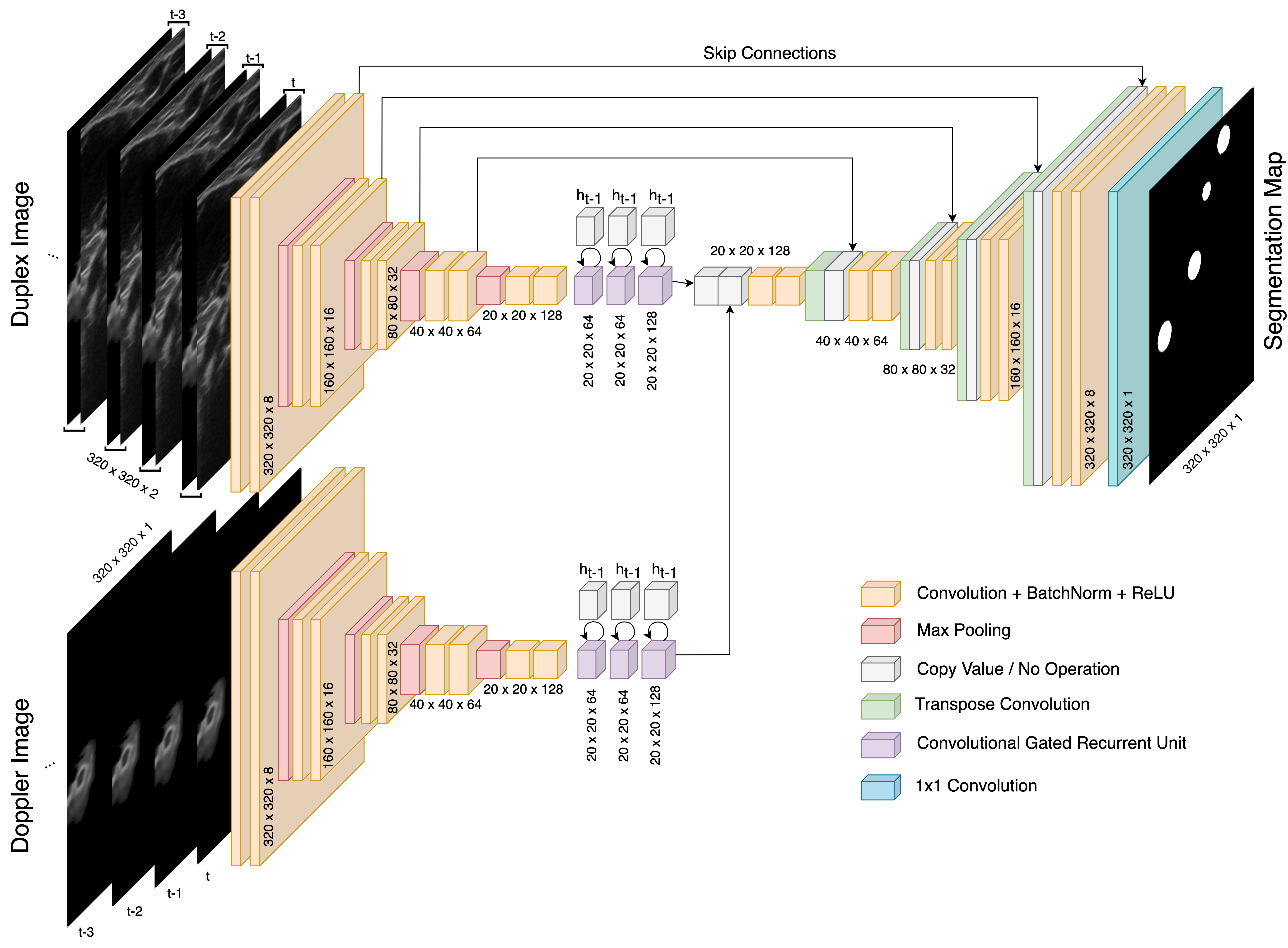}
\caption{The architecture of the proposed DopUS-Net.}
\label{fig:dop_us_net}%
\end{figure*}

\par
Benefiting from the development of deep learning, neural networks have been widely employed to solve real-time image segmentation tasks. In the field of medical image segmentation, U-Net~\cite{ronneberger2015u} is one of the most commonly used and successful networks. The U-Net and its variants have also been used for US vessel segmentation tasks~\cite{chen2020deep, jiang2021autonomous}.

\par
In this work, the proposed DopUS-Net also uses the classic U-shape backbone for the encoder and decoder (see Fig.~\ref{fig:dop_us_net}). Contrary to other networks which also take the Doppler inputs into account~\cite{jiang2021automatic_baichuan, chen2020deep}, the DopUS-Net uses two separate encoders. The paired duplex image is considered as a two-channel input of the top encoder, while the bottom encoder only uses the pre-processed color Doppler images. Since only Doppler images are used in the bottom encoder, the network is forced to filter out the noise and focus on the main artifacts. This structure can limit the negative impact of the unstable Doppler signal (e.g., spatial accuracy and consistency), but still preserves the functionality of a region proposal signal. The skip connections between the top encoder and decoder are used to ensure the spatial correctness of the B-mode image meanwhile also including the region proposal functionality of the Doppler images, respectively. This extensive use of the Doppler signal in both encoders proved to be beneficial in terms of performance, which also indicates its usefulness. Besides, DopUS-Net employs batch normalization instead of dropout between the layers as suggested by~\cite{garbin2020dropout}, where it is claimed that using dropout can have detrimental effects on CNN training. 


\par
Due to the continuity of vascular tissues, \gls{rnn} structures are employed to facilitate the segmentation task by taking advantage of sequential historic information. Due to the weight update rule, gradient exploding or vanishing can happen when working with long input sequences~\cite{bengio1994learning}. To tackle this problem, \gls{lstm}~\cite{hochreiter1997long} and \gls{gru}~\cite{cho2014learning} networks have been developed by using multiple gates internally to keep track of important information meanwhile discarding irrelevant data. Chung~\emph{et al.} tested both \gls{rnn}s and claimed that \gls{gru} networks are less complex while maintaining comparable performance to an \gls{lstm}~\cite{chungEmpiricalEvaluationGated2014}. Thereby, \gls{gru}s are used in the proposed network structure, which leads to fewer computations and therefore better inference time. To further reduce the number of parameters needed, Ballas~\emph{et al.} introduced the \gls{convGRU} network~\cite{ballas2015delving}. The hidden state $\textbf{h}_t$ extracted from sequential inputs is computed as follows: 
\begin{equation} \label{eq:convgru}
	\begin{split}
		\textbf{z}_{t} & = \sigma(\textbf{W}_{z} [\textbf{x}_{t}, \textbf{h}_{t-1}]^T) \\
		\textbf{r}_{t} & = \sigma(\textbf{W}_{r}  [\textbf{x}_{t}, \textbf{h}_{t-1}]^T) \\
		\tilde{\textbf{h}}_{t} & = \tanh(\textbf{W} [\textbf{x}_{t}, (\textbf{r}_{t} \odot \textbf{h}_{t-1})]^T) \\
		\textbf{h}_{t} & = (1-\textbf{z}_{t})\textbf{h}_{t-1}+\textbf{z}_{t} \tilde{\textbf{h}}_{t} \\
	\end{split}
\end{equation}
where $\odot$ is an element-wise multiplication, $\textbf{x}_t$ is the sequential input, $\textbf{h}_{t-1}$ is the hidden state computed in the last iteration, $\textbf{z}_{t}$ is an update gate that dominates the degree to which the unit refreshes its content, and $\textbf{r}_{t}$ represents the reset gate. $\textbf{W}_{z}$ and $\textbf{W}_{r}$ are the weights for the update gate and reset gate, respectively. $\tilde{\textbf{h}}_{t}$ further operates on combining the input with an updated hidden state of the previous time step. The final hidden state $\textbf{h}_t$ is computed based on the result of the update gate.

\par
The~\gls{convGRU} is used at the bottleneck in both encoders to enable state tracking functionality. The bottom encoder is, therefore, able to filter out noise and speckles from the Doppler effect meanwhile maintaining a representative view of the recurring pulsation of the Doppler artifacts in the arteries. The top encoder can use its historic information to emphasize previously found vessel locations in the current segmentation. The fusion of the two encoders and their respective recurrent unit is done via a concatenation. Afterward, a convolution ties the input together and reduces the feature dimensionality. Consequently, the result of the combined bottleneck is fed into the decoder where a $1\times1$ convolution is used at the initial image resolution to generate the binary output mask (see Fig.~\ref{fig:dop_us_net}).


\subsection{Training}
\subsubsection{Data Augmentation}
\par
In this work, data augmentation is applied during the training process to improve robustness and reduce potential overfitting~\cite{shorten2019survey}. To maintain a constant augmentation pattern for the complete sequence, the same parameters were applied to all images in the same sequence. The spatial augmentations parameters are listed in TABLE~\ref{tab:augmentations}. During the training, a random number is chosen among the given ranges. We empirically determined that applying horizontal and vertical shifts of only $\pm1\%$ was optimal for our model's performance. Larger shifts, in combination with the other parameters, decreased the model's efficacy on real US data, although further optimization is possible in future work.

	\begin{table}[htb!]
		\centering
		\caption{Applied augmentations to the image sequence}%
		\label{tab:augmentations}
		\begin{tabular}{l|c}%
		\noalign{\hrule height 1.2 pt}
			 \bfseries{Augmentation Method}                                          & \bfseries{Range / Value} \\
			 \hline
			 Horizontal Shift  & [-1\% , 1\%] \\
			 Vertical Shift  & [-1\% , 1\%] \\
			 Rotation & [-15$^\circ$, 15$^\circ$] \\
			 Scale & [80\%, 120\%] \\
			 Horizontal Flip Probability & 50\% \\
		\noalign{\hrule height 1.2 pt}
		\end{tabular}
	\end{table}

\subsubsection{Loss Function}
\par
In medical binary segmentation problems,~\gls{bce} is the most commonly used loss function~\cite{bertels2019optimizing}. The pixel-wise approach of class probabilities leads to a smooth gradient curve. However,~\gls{bce} is sensitive to imbalanced classes. Particularly in our case, the limb arteries are very small in comparison to the background. The area recorded by the US probe is around $2300~mm^2$ (transducer length: $52~mm$, depth: $45~mm$), while the radial or the ulnar artery only covers approx $5-10~mm^2$ (diameter: $2-2.5~mm$~\cite{beniwal2014size}) of this region. This leads to a possible class imbalance of $1:250$. To tackle this problem, the soft dice loss function~\cite{sudre2017generalised} is used instead in this work. The soft dice loss for all classes is then averaged to obtain a final score. Besides, Bertels~\emph{et al.} also reported that the soft dice loss is recommended when using the dice score as a performance index~\cite{bertels2019optimizing}. The computation of the soft dice loss $L$ is presented in Eq.~(\ref{eq:sdl}). 

\begin{equation} \label{eq:sdl}
	L =  1- \frac{2 | \tilde{Y}  \cap Y | +1}{ | \tilde{Y} | + | Y | + 1} = 1- \frac{2 * \sum_{i=1}^{N} \tilde{y} * y + 1}{ \sum_{i=1}^{N} \tilde{y} ^2 + \sum_{i=1}^{N} y^2 + 1} 
\end{equation}
where $N$ represents the total amount of pixels, $y$ and $\tilde{y}$ symbolize the target value (ground truth) and predicted label value, respectively.

\subsubsection{Training Details}
\par
To optimize the weights of DopUS network, the Adam~\cite{kingma2014adam} optimizer is used in this work. Additionally, mixed-precision training~\cite{micikevicius2018mixed} is enabled to speed up the training process. This form of auto casting weight precision reduces the overall memory requirements and therefore increases the training speed significantly. There are $20$ images in each sequence data set. Since the robot moved at a pace of $10~mm/s$ and the duplex images are recorded in $10~fps$, the length of each sequence is around $20~mm$. The parameters for the training epoch and the batch size are set to $1000$ and $16$, respectively. The initial learning rate is 1e-4 and halved every $250$ iterations. The use of a small learning rate at the end of the training can benefit the accurate convergence. To prevent unnecessary training epochs, an early stopping mechanism was implemented. If the validation loss is not decreasing within $15$ epochs, the algorithm is assumed to have converged and no further improvements are expected. Since recurrent structures are present in the network, truncated backpropagation through time is applied. Every $4$ steps the weights of the network are updated. This value represented a good compromise between performance and training time.

\section{Robotic US Scanning and Artery Re-Identification}
\par
To guarantee the patient's safety and image quality, the robotic scans are performed using compliant control~\cite{jiang2021autonomous}. The scan trajectory is manually defined by selecting the start and end position on the patient's skin. To achieve a highly accurate 3D visual representation of the artery tree, Doppler images and B-mode images are used jointly to extract the vessel from the cross-sectional images. A stable and continuous Doppler signal is beneficial for the performance of the proposed DopUS-Net. To account for unstable Doppler signals and problems introduced by the blood pulsation, a tracking algorithm is developed in Section~\ref{sec:doppler_tracker}. Based on this tracker, an online image check is activated to guarantee sufficient quality of the Doppler signal. Once this check identifies poor Doppler signal, which is often accompanied by unstable segmentation results, a re-identification procedure is performed to relocate the Doppler signals (see Section~\ref{sec:reident_procedure}).



\subsection{Compliant Control Architecture}~\label{sec:control}
\par

The impedance controller is often used to maintain the contact force between the probe and the contact surface~\cite{jiang2021autonomous,hennersperger2016towards}. Due to the use of built-in joint torque sensors in all seven joints, the impedance control law can be defined as follows:

\begin{equation}\label{eq_impedance_law}
\tau = \textbf{J}^{T}[\textbf{F}_d + \textbf{K}_m e + \textbf{D} \dot{e} + \textbf{M} \Ddot{e}]
\end{equation}
where $\tau$ is the computed torque, $J^{T}$ is the transposed Jacobian matrix, $e = (x_d - x_c)$ is the pose error (position and orientation) between the current pose $x_c$ and the target pose $x_d$ in Cartesian space, $\textbf{F}_d$ is the supposed exerted force/torque at end-effector, $\textbf{K}_m$, $\textbf{D}$ and $\textbf{M}$ represent the matrices of stiffness, damping and inertia terms, respectively. According to~\cite{hennersperger2016towards}, the stiffness in the direction of the probe centerline is usually set in the range $[125, 500]~N/m$ for human tissues.

\subsection{Doppler Signal Tracker}~\label{sec:doppler_tracker}
\par
Due to the heart pulsation, the Doppler images are not as stable as B-mode images. Regarding the scanning of limb arteries, both the accuracy and strength of Doppler signals are influenced by factors like the contact condition and the probe orientation. This may impair the performance of the segmentation. However, the ability to automatically identify the flow location in the US image is important, particularly for developing autonomous scanning programs. The inclusion of the Doppler signal enables quality-aware robotic screening, which can assess the segmentation results online and leads to improved 3D reconstruction results across the scans.  

\par
To assess the quality of real-time Doppler signals, we monitor the centers of detected flow areas [see Fig.~\ref{fig:pre_processing_pipeline}~(c)]. Due to the continuity of the objects, the corresponding centers should be close to each other on consecutive frames. To compute the centers, all pre-processed Doppler images are examined for contours using the minimum enclosing circle algorithm, which is a  standard algorithm wrapped in the open-source library OpenCV~\cite{bradskiOpenCVLibrary2000}. Considering the inevitable Doppler noise, only the contours above a certain empirical threshold (radius $> 1.2~mm$) are kept. Afterward, the centers are assigned to tracking objects capturing the historical information of the Doppler signal. To further process the potential case that multiple vessels are displayed on a single image, the centers need to be assigned to an existing tracking object $\mathbb{O}^{t-1}\in R^{N\times1}$ or to be considered as a new object, where $N$ is the number of the tracked objects, namely blood vessels. To this end, the distance between the centers extracted from the images recorded at time $t-1$, $\textbf{C}^{t-1}\in R^{N\times2}$, and the centers computed on the current images at time $t$, $\textbf{C}^{t}\in R^{N'\times2}$, is computed as follows:

\begin{equation}~ \label{eq:distance}
    \textbf{D}_i^t =  \text{CalDis(} \textbf{C}_{i}^{t-1},~\textbf{C}_{all}^{t}) 
\end{equation}
where $\text{CalDis()}$ represents the operation to compute Euclidean distances between one point and each element in a point set. $i = 1, 2,..., N$ is the iterator referring to the centers of tracked objects, $\textbf{C}_i^{t-1}\in R^{1\times2}$ is the $i$-th center saved at $\mathbb{O}^{t-1}$, $\textbf{C}_{all}^t\in R^{N'\times2}$ is the set of centers computed based on the Doppler image obtained at time $t$, $N'$ is the filtered number of contours present at time $t$, and $\textbf{D}_i^t\in R^{N'\times1}$ is the distances between $\textbf{C}_i^{t-1}$ and all elements of $\textbf{C}_{all}^t$. The minimum distance $d_{min}$ is stored with its associated index $j$ and used to update the tracking object $\mathbb{O}^{t-1}$ as described in Algorithm~\ref{algorithm_re_identification}.


\par
The tracking object $\mathbb{O}$ is updated based on the distance between the contours' centers calculated in the last frame $t-1$ and the current frame $t$ [Eq.~(\ref{eq:distance})]. The center point of the current frame with the smallest distance to a respective tracking object's center point, $\textbf{C}_{all}^t(j)$,  will become its new center point. To avoid arbitrary center point assignment, a distance limit $T_d$ of maximal $30$ pixels is empirically set, which means $4.2~mm$ maximum deviation from the previous center point. For all center points that could not be matched to an existing tracking object, a new tracking object is created. If, on the other hand, no Doppler signal could be assigned to a tracking object, the center point of the previous frame is stored. Additionally, this value will be labeled so that a distinction between a tracked value and a copied value can be made.

\begin{algorithm}[htb] 
\caption{Doppler Signal Tracker}\label{algorithm_re_identification}
\KwIn{previous tracking object $\mathbb{O}^{t-1}$, extracted vessel centers on previous and current images $\textbf{C}_{all}^{t-1}$ and $\textbf{C}_{all}^t$}
\KwOut{current tracking object $\mathbb{O}^{t}$}
$\mathbb{O}^{t}~\xleftarrow~\mathbb{O}^{t-1}$ \;
\For{$i=1$; $i\leq \text{len}(\mathbb{O}^{t-1})$; $i++$}
{
    $\textbf{D}_i^t$ $~\xleftarrow~$ $\text{CalDis(} \textbf{C}_{i}^{t-1},~\textbf{C}_{all}^{t})$ [Eq.~(\ref{eq:distance})]\;
    $d_{min}~\xleftarrow~\text{min}(\textbf{D}_i^t)$ = $\textbf{D}_i^t$(j)\;
    \eIf{$d_{min} \leq T_d$}
    {
        $\mathbb{O}^{t}(i)~\xleftarrow~\textbf{C}_{all}^t(j)$ \;
    }
    {
        $\mathbb{O}^{t}$ $~\xleftarrow~$ [$\mathbb{O}^{t}$, $\textbf{C}_{all}^t(j)$]\;
    }
}
\end{algorithm}

\subsection{Artery Location Re-Identification} \label{sec:reident_procedure}
\par
\begin{figure}[ht!]
\centering
\includegraphics[width=0.45\textwidth]{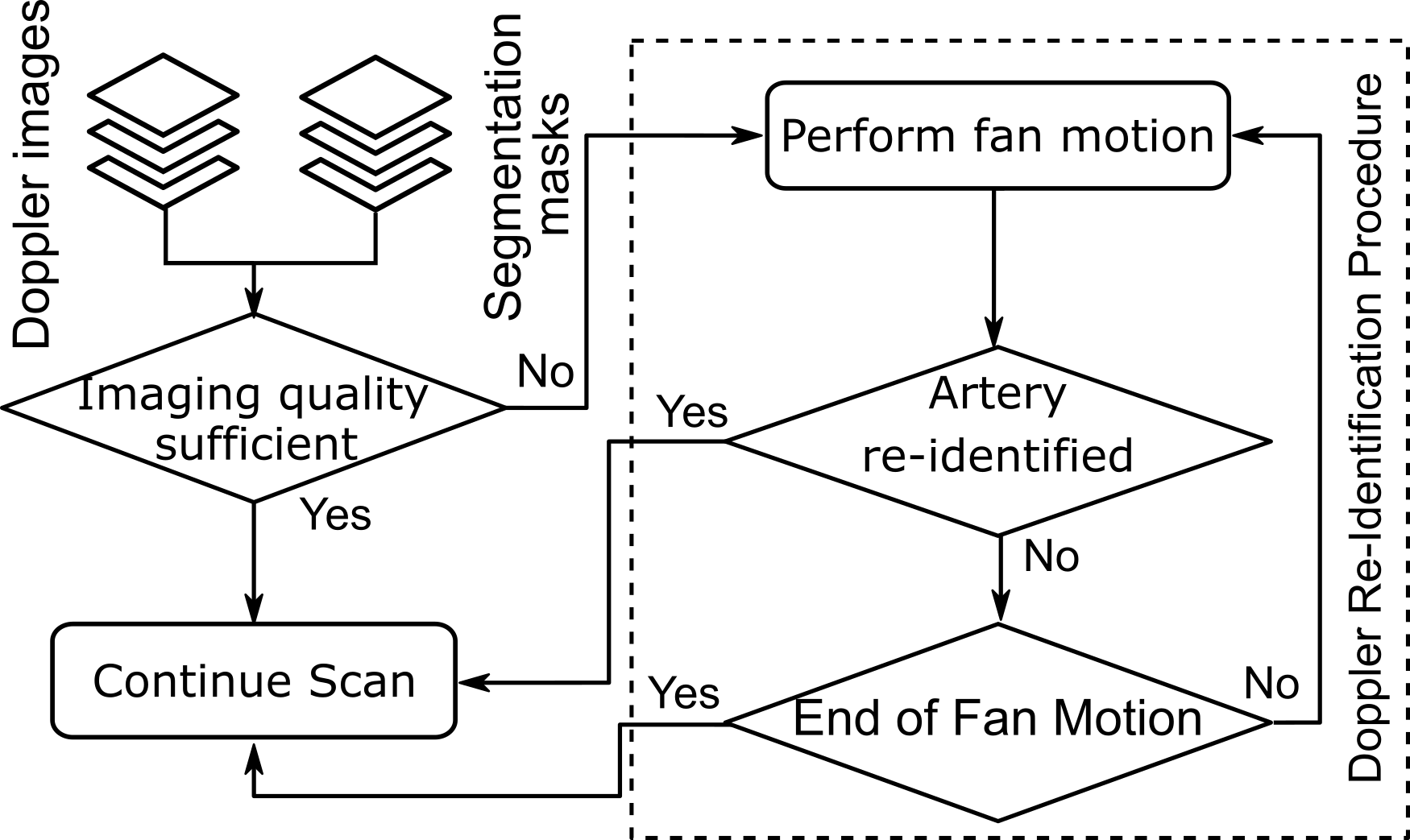}
\caption{The flow chart of the artery re-identification procedure.}
\label{fig:reident_flow}
\end{figure}

Compared with existing vessel segmentation and tracking approaches, the use of the Doppler signal enables quality-aware robotic scanning, leading to robust and accurate 3D visualization. Although some learning-based approaches have been proposed to achieve accurate segmentation, there is no approach that can evaluate the segmentation results online. Regarding pure segmentation tasks from images, a large segmentation error on a few slices would not affect the overall performance. However, such errors may lead to incorrect robotic motion when the segmentation is further used to control a robotic manipulator. Thereby, it is necessary to re-identify the artery location once the segmentation quality is inadequate. Here we consider imaging quality to be insufficient when the real-time segmentation results are not consistent with the Doppler images. To ensure a consistently high segmentation performance, a re-identification procedure is developed (see Fig.~\ref{fig:reident_flow}). 

\par
To evaluate the segmentation performance, Doppler images and real-time segmentation results are jointly used as inputs. The imaging quality is determined by comparing the last center point of the tracking object $\mathbb{O}$ to the segmentation mask. A tracking object is only considered if at least $20\%$ of its center points within the last $15$ frames represent new values (not a copy from the last one). The threshold is determined based on empirical studies and the assumption that a flow is visible in the last $15$ frames. Considering that the images are recorded in $10~fps$, $15$ frames correspond to a time window of $1.5~s$ Note that the interval is based on the assumption that at least one heartbeat can occur within that time frame and may need to be adjusted for patients with slower heart rates.

\par
\begin{figure}[ht!]
\centering
\includegraphics[width=0.40\textwidth]{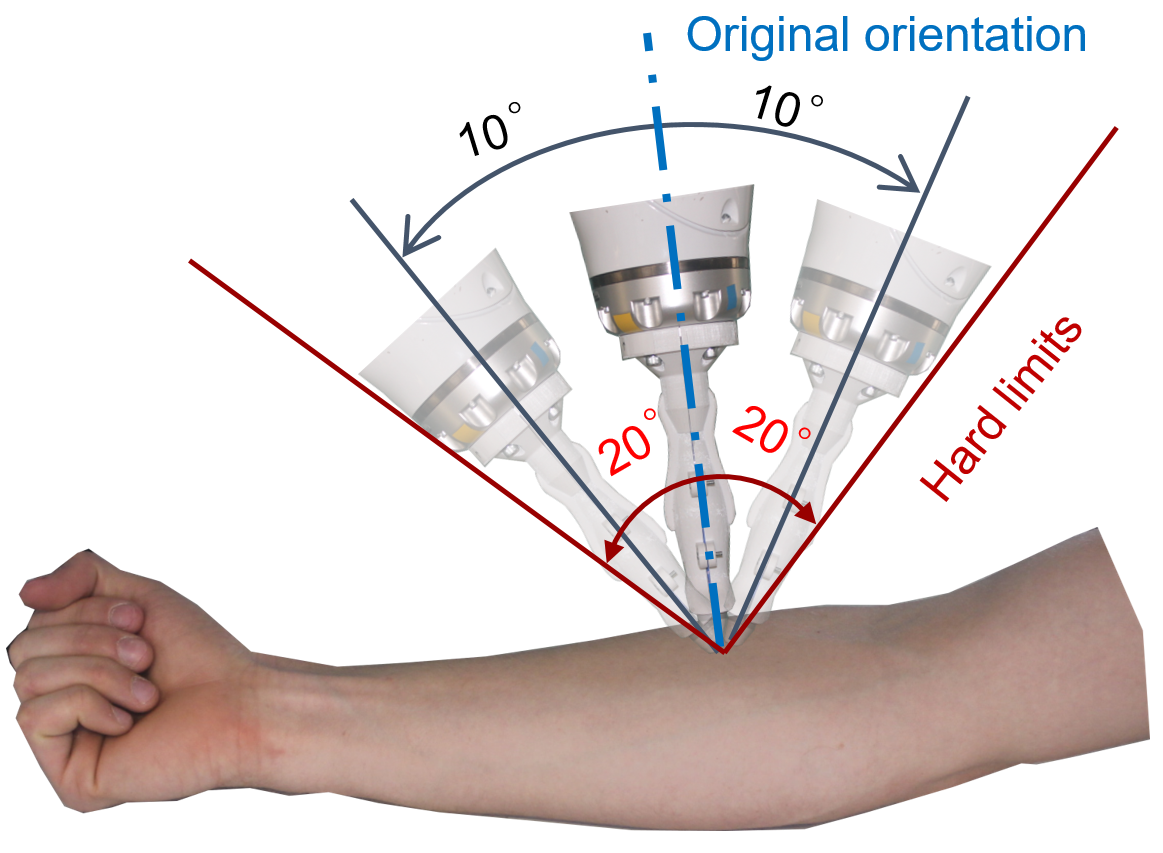}
\caption{The illustration of the out-of-plane rotation for the re-identification process.}
\label{fig:reident_vis}
\end{figure}

Furthermore, if at least one of the tracked center points of the valid tracking objects is within the areas of the predicted vessels, the quality is assumed to be sufficient. Otherwise, we consider that the tracker and real-time segmentation results are in conflict and thereby automatically start the re-identification algorithm. In this case, the current probe pose is referred to as the original orientation for this re-identification procedure (see Fig.~\ref{fig:reident_vis}). Due to the physical principles of the Doppler signal, its quality gets better when the beam of the US probe is aligned with the direction of the vessel flow. Better Doppler imaging quality can further facilitate the segmentation performance of the proposed DopUS network. Thereby, once the system is aware that the imaging quality is not good enough, an out-of-plane fan motion is performed to obtain better imaging quality (see Fig.~\ref{fig:reident_vis}). However, the contrast of the B-mode images decreases when tilting the probe away from the optimal perpendicular pose. In order to keep both the US and Doppler image quality as high as possible, the fan motion should only rotate as far as necessary.

\par
Taking that into account, the probe is rotated in the out-of-plane direction in $5^\circ$ steps every time. In total, a range of $\pm10^\circ$ from the current probe orientation can be covered. After every step, the robot stops for two seconds to be notified if the Doppler quality check reports a good signal. A maximum of five different orientations (out-of-plane rotation deviation from the current pose: [$-10^{\circ}$, $-5^{\circ}$, $0$, $5^{\circ}$, $10^{\circ}$]) are visited, assuming that no sufficient quality of the Doppler signal is detected. In this case, the robot returns to its original orientation and continues the sweep. In order to move on to a potentially better location for the Doppler signal the re-identification process is blocked for at least three seconds. If on the other hand the Doppler signal relocalizes the vessel during the re-identification process, the current orientation is used to update the trajectory orientation and the sweep is continued. Finally, to prevent the robot from moving too close to the human arm, a security threshold referring to the original direction is implemented. So the out-of-plane rotation is stopped if it deviates more than $20^\circ$ compared to the original orientation.


\section{Results}
\subsection{Experimental Setup}
\par
\begin{figure}[ht!]
\centering
\includegraphics[width=0.40\textwidth]{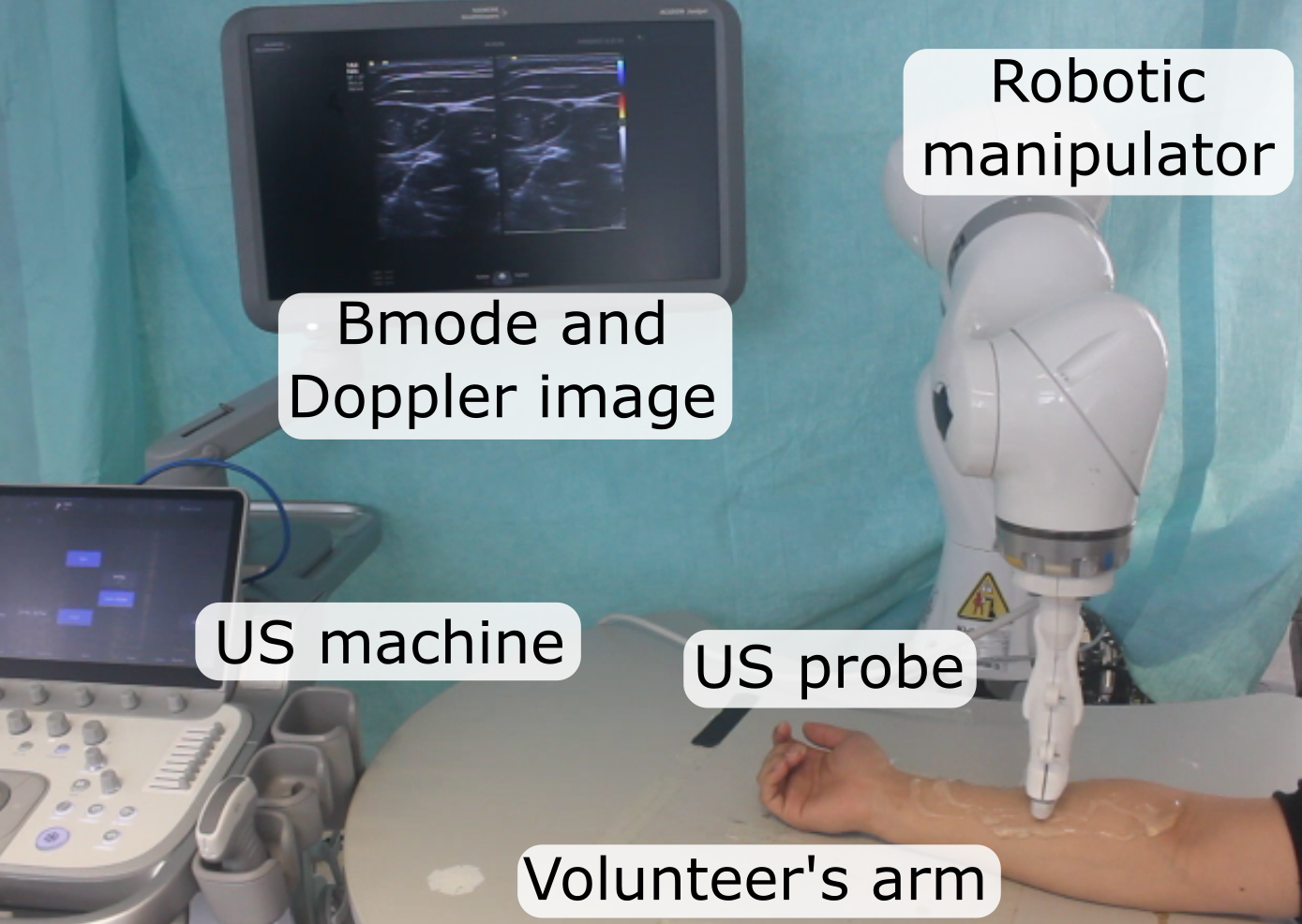}
\caption{The experimental setup: robot manipulator with attached linear US probe and the used US machine.}
\label{fig:robot_setup}
\end{figure}

The overall setup is visualized in Fig.~\ref{fig:robot_setup}. A linear probe is rigidly attached to the robot manipulator (LBR iiwa 14 R820, KUKA GmbH, Germany). The connector between the transducer and the end-effector of the robotic arm is a custom-designed probe holder. 
To guarantee safety and imaging quality, the robotic arm is controlled using impedance control during scans (see Section~\ref{sec:control}). The desired force in the probe centerline is set to $1~N$, and the stiffness is $200~N/m$. The main objective of using a non-zero force is to ensure firm contact between the probe and patients during scans, while force could be varied from $1~N$. However, during the re-identification process, the control mode automatically switches to a position control strategy to maintain the same position when restarting.

\par
Regarding the image processing part, the duplex images are recorded and fed to the DopUS-Net for real-time vessel segmentation. The inference time of DopUS-Net, including pre-processing and the online imaging quality evaluation process, is around $20~ms$ on average, which is quick enough to process the real-time US images captured at $10~fps$ ($100~ms$). In addition, the segmentation masks are stacked in 3D based on the robotic tracking information and visualized in real-time on a visualization platform (ImFusion Suite, ImFusion GmbH, Germany).

\subsection{Segmentation Performance on Volunteers}
\par

\begin{table*}[t]
\begin{center}
\caption{Comparison of the results}
\label{tab:segmentation_results}
\resizebox{0.98\textwidth}{!}{
\begin{tabular}{cccccccccccc}
\noalign{\hrule height 1.2 pt}
 Network & \makecell{Top\\Encoder} & \makecell{Bottom\\Encoder} & \makecell{\#\\Parameters} & \makecell{\\0} & \makecell{\\1} & \makecell{\\2} & \makecell{Patient\\3} & \makecell{\\4} & \makecell{\\5} & \makecell{\\6} & \makecell{Dice Score\\Mean (SD)} \\
\hline
U-Net & B & - & 0.6 M &  0.61 & 0.51 & 0.45 & 0.40 & 0.45 & 0.29 & 0.38 & 0.44 (0.09) \\
U-Net & BD & - & 0.6 M & 0.83 & 0.59 & 0.54 & 0.52 & 0.49 & 0.39 & 0.45 & 0.55 (0.13) \\
DopUS-Net\textsuperscript{(0)} & BD &  D & 1.3 M & 0.83 & 0.61 & 0.58 & 0.60 & 0.54 & 0.43 & 0.47 & 0.58 (0.12) \\
\hline
VesNet & BD-RNN & - & 2.6 M & 0.80 & 0.62 & 0.60 & 0.43 & 0.33 & 0.42 & 0.43 & 0.52 (0.15) \\
VesNet+ & BD-RNN & - & 6.3 M & 0.80 & 0.62 & 0.68 & 0.58 & 0.58 & 0.47 & 0.53 & 0.61 (0.10) \\
U-Net & BD-RNN & - & 3.0 M & 0.86 & 0.61 & 0.71 & 0.66 & 0.68 & 0.60 & 0.50 & 0.66 (0.10) \\
U-Net+ & BD-RNN & - &  6.5 M & 0.78 & 0.58 & 0.37 & 0.40 & 0.36 & 0.32 & 0.41 & 0.46 (0.15) \\
\hline
DopUS-Net\textsuperscript{(1)} & B-RNN & D-RNN{$^{\clubsuit}$} & 6.2 M & 0.78 & 0.54 & 0.45 & 0.56 & 0.55 & 0.27 & 0.18 & 0.48 (0.19) \\
DopUS-Net\textsuperscript{(2)} & BD & D-RNN & 3.7 M & 0.87 & 0.69 & 0.76 & 0.76 & 0.72 & 0.56 & 0.61 & 0.71 (0.10) \\
DopUS-Net\textsuperscript{(3)} & B-RNN & D-RNN & 6.1 M & 0.88 & \textbf{0.71} & 0.79 & 0.76 & 0.75 & 0.57 & \textbf{0.61} & 0.72 (0.10) \\
DopUS-Net\textsuperscript{(4)} & BD-RNN &  D-RNN & 6.1 M & \textbf{0.88} & 0.69 & \textbf{0.79} & \textbf{0.78} & \textbf{0.76} & \textbf{0.62} & 0.60 & \textbf{0.73 (0.09)} \\
\noalign{\hrule height 1.2 pt}
\end{tabular}
}
\end{center}
{\textit{*Nomenclature:}
B:B-Mode, D: Doppler, RNN: convGRU module, +: increased parameters, DopUS-Net\textsuperscript{(x)}: specific DopUS-Net version, $^{\clubsuit}$: additional skip connections from the bottom encoder to the decoder.}
\end{table*}

\par
The network was trained on a workstation (GPU: GeForce GTX Titan X, CPU: Intel i7-4820K). To measure the similarity between the manually annotated ground truth data $Y$ (see Section \ref{sec:dataset}) and the binary segmentation result $\tilde{Y}$, the Dice-Score $C_{dice}$ is applied (Eq.~\ref{eq:dice_score}).

\begin{equation} \label{eq:dice_score}
	\begin{split}
		C_{dice} & =  \frac{2 | \tilde{Y}  \cap Y |}{ | \tilde{Y} | + | Y |}\\
	\end{split}
\end{equation}

\par
We employed the leave-one-out-cross-validation (LOOCV) method [44] to optimize the performance of the DopUS network with a limited training dataset. LOOCV entails training the model on data from six patients and validating it on the remaining one, iteratively for each patient. This reduces the potential bias of the model compared with conventional approaches that split the data into fixed training and test sets. LOOCV ensures that the model is evaluated on data from all seven patients, albeit at a higher computational cost. It yields a more robust and reliable estimate of the network performance.

\par
To compare the performance with existing networks and further explore the most effective way to incorporate the Doppler signal, the performances of different network architectures have been investigated. All presented networks are implemented using PyTorch\footnote{https://pytorch.org/}, and they are trained from scratch based on the same data set in each case. The results have been summarised in TABLE~\ref{tab:segmentation_results}. To ensure a valid comparison throughout the study, all the models are trained using the same configuration on the same dataset. The baseline and comparison networks represented are U-Net~\cite{ronneberger2015u} and VesNetSCT++~\cite{jiang2021automatic_baichuan}. \textbf{It must be noted}, that the VesNetSCT++ architecture is a custom re-implementation as the source code was not publicly available. Results should be treated with care. The self-developed VesNetSCT++ is carefully implemented by following the architecture presented in~\cite{jiang2021automatic_baichuan}. The final version achieved comparable dice scores on small blood vessels (diameter is $2\mbox{-}3~mm$) with respect to the original paper \cite{jiang2021automatic_baichuan}. To maintain simplicity, the VesNetSCT++~\cite{jiang2021automatic_baichuan} is referred to as VesNet in the remaining part of this work.

\par
As all involved network architectures use the U-shape backbone, descriptions of their top and bottom encoders are used to facilitate the understanding of the key differentiation aspects. The top encoder represents the encoder in the U-Net/VesNet architecture. Since the proposed DopUS-Net follows the approach of two separate encoders, the second one is described as the bottom encoder (see~ Fig.~\ref{fig:dop_us_net}). The corresponding inputs of these two encoders used in different networks are specified in TABLE~\ref{tab:segmentation_results}. The letter ``B" and ``D" are B-mode and Doppler images, respectively. BD represents a two-channel tensor consisting of the combined duplex images. RNN corresponds to the convGRU module, which is used before the bottleneck at the highest feature dimensionality space.

\subsubsection{Effectiveness of Doppler Signal on Vessel Segmentation}
\par
To validate whether the Doppler signal can enhance the segmentation accuracy from US images, the classic U-Net~\cite{ronneberger2015u} is used as a baseline in this study. In the first two rows of TABLE~\ref{tab:segmentation_results}, we compare the effect of pure B-Mode image input to a two-channel Doppler and B-Mode tensor input. The average segmentation results are improved from $0.44$ to $0.55$ in terms of dice score. Such improvement is consistent with the results reported by B. Jiang~\emph{et al.}~ using VesNet~\cite{jiang2021automatic_baichuan}. Then, the additional bottom encoder using the Doppler images as input is tested. The double encoder approach further improved the segmentation results to $0.58$. It is noteworthy that DopUS-Net\textsuperscript{0} consists of a classic U-Net and an additional encoder with the same architecture as the upper encoder. Thereby, we consider the use of Doppler signals can significantly improve the segmentation performance.


\subsubsection{Effectiveness of RNN on Vessel Segmentation}
\par
Due to the continuity of blood vessels, RNNs are used to take advantage of historical information in this study. To evaluate the impact of adding an RNN, we compare the performance of the U-Net with and without the RNN module. Based on the experimental results in TABLE~\ref{tab:segmentation_results}, the performance is significantly improved from $0.55$ ($2$-nd row) to $0.66$ ($6$-th row) when the RNN module is used. Similarly, the best performing version of the DopUS-Net\textsuperscript{(4)} with RNN for both encoders increases its performance by roughly $15\%$. Thereby, we consider the RNN to be beneficial for the achievement of accurate vessel segmentation. 

\par
To account for parameter differences introduced by the different structures, the network parameters of the comparison networks have been increased by increasing the number of filters in each layer. The version with increased parameters of U-Net and VesNet are marked as U-Net+ and VesNet+, respectively. The parameter numbers of U-Net+ ($6.3~M$) and VesNet+ ($6.5~M$) roughly match the size of the final DopUS-Net ($6.1~M$). Based on the experimental results, the VesNet+ achieved a better result ($0.61$), while the U-Net+ performed significantly worse than the baseline (only $0.46$). We consider this \final{to} be caused by the overfitting when the number of parameters (U-Net) significantly increased. To address this issue, necessary fine-tuning and regularization would be required, e.g., dropout, which in turn would result in a parameter reduction. Therefore, the better performing version with fewer parameters is used for further comparison.

\subsubsection{Effectiveness of Network Structures on Vessel Segmentation}
\par
To further explore the most effective way of incorporating the Doppler signals to facilitate segmentation accuracy, four different versions of DopUS-Net have been investigated. DopUS-Net\textsuperscript{(1)} and DopUS-Net\textsuperscript{(3)} only use the Doppler signal in the bottom encoder, while DopUS-Net\textsuperscript{(2)} and DopUS-Net\textsuperscript{(4)} use Doppler signals in both encoders. First, we quantitatively evaluated the effect of the skip connections. DopUS-Net\textsuperscript{(1)} incorporates additional skip connections from the bottom encoder to the decoder. Compared to DopUS-Net\textsuperscript{(3)}, which only has the skip connections from the top encoder to the decoder, the performance of DopUS-Net\textsuperscript{(1)} is significantly reduced, deteriorating the dice score by $0.24$. We consider that this is because of the inherent characteristic instability of the Doppler signal. The results affirm the assumption that the Doppler signal can be seen as a good region proposal without spatial accurateness. \final{In addition}, compared to the final version DopUS-Net\textsuperscript{(4)}, DopUS-Net\textsuperscript{(2)} and DopUS-Net\textsuperscript{(3)} remove the RNN module and Doppler signal, respectively, from the top encoder. Although the experimental result demonstrated that DopUS-Net\textsuperscript{(4)} achieved better performance than DopUS-Net\textsuperscript{(2)} and DopUS-Net\textsuperscript{(3)}, the improvement is minor. All of these three networks significantly improve the state-of-the-art results (U-Net: $0.66$, VesNet: $0.61$) to $0.71$, $0.72$ and $0.73$. Overall the insight can be taken, that the second encoder forces the network to emphasize on the Doppler effect. The results show that the intuitive inclusion and emphasized weight on the Doppler signal is justified in practice. More segmentation results of live US frames can be found in the online \textbf{video}.
\subsection{Validation of the Artery Re-Identification Process}

\begin{figure*}[ht!]
\centering
\includegraphics[width=0.95\textwidth]{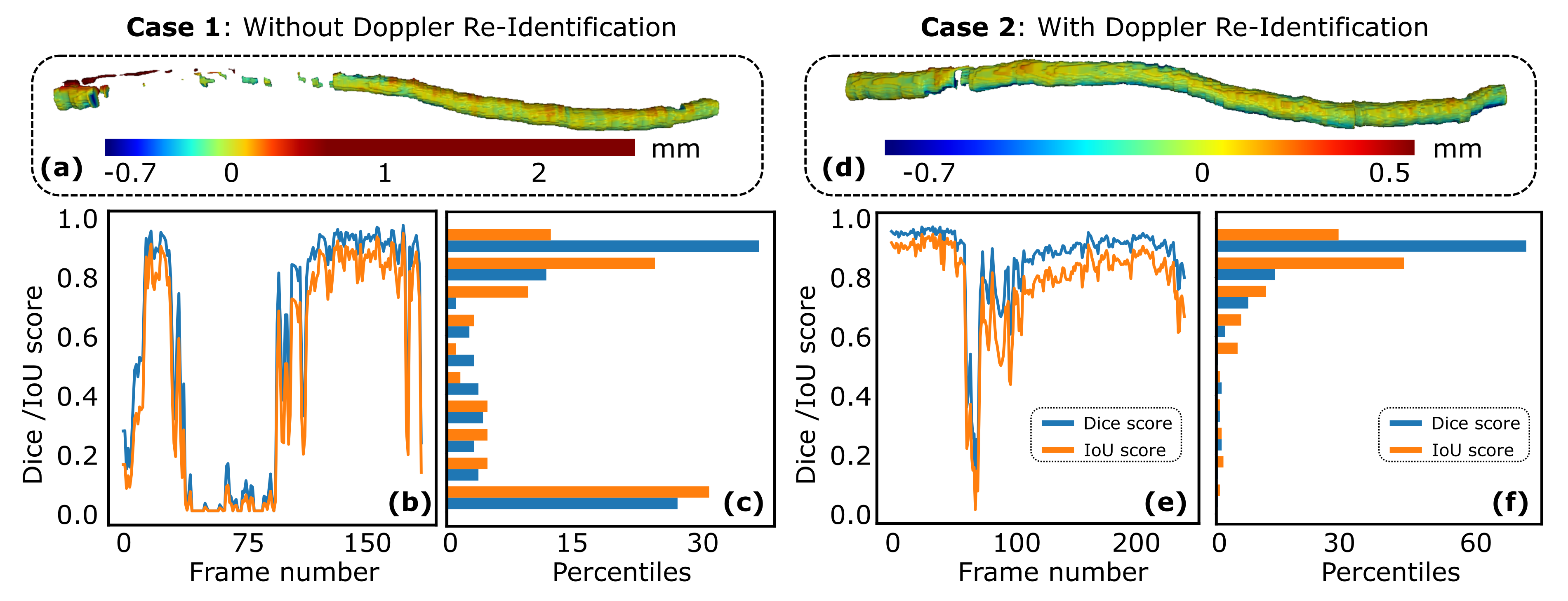}
\caption{A representative comparison between the robotic scans with and without artery location re-identification process. (a) is the heatmap of the geometry difference between the automatic segmentation results and the ground truth. (b) and (c) are the dynamic and statistically summarized Dice/IoU scores, respectively. (d), (e) and (f) are the results obtained when scanning with enabled re-identification process.}
\label{fig:vessel_evaluation}%
\end{figure*}


\par
To further validate whether the proposed artery location re-identification approach can improve the robustness and final 3D compounding results, experiments were carried out on various volunteers. To quantitatively compare the difference between the scans with and without the re-identification process, two robotic scans with the same initial scanning trajectory were performed on the same volunteer. The initial trajectory was manually given by selecting the start and end points. The ulnar artery on the arm was selected as the anatomy of interest. Representative results of two scans with and without the re-identification process, respectively, on the same volunteer, are depicted in Fig.~\ref{fig:vessel_evaluation}. To avoid misleading information, it is noteworthy that the segmentation noise of the DopUS-Net has been removed to focus on the vessel of interest. The heatmap intuitively shows the geometric offset between the segmentation result of the DopUS-Net and the manually annotated ground truth. Similar to the dataset used for training and evaluation of the proposed network structure, the ground truth annotation for the artery re-identification validation was performed under the supervision of an experienced sonographer.

\begin{table}[!ht]
\centering
\caption{Segmentation results over volume [Mean (SD)]}
\label{tab:seg_eval_metrics}
\begin{tabular}{ccc}
\noalign{\hrule height 1.2 pt}
Re-Identification                         & Dice Score & IoU    \\
\hline
Enabled &  $0.86~(0.14)$   &  $0.78~(0.16)$ \\
Disabled   &  $0.54~(0.39)$   &  $0.47~(0.37)$  \\
\noalign{\hrule height 1.2 pt}
\end{tabular}
\end{table}

\begin{table*}[tb]
\centering
\caption{Overall Performance of the Robotic Screening Framework [Mean(SD)]}
\label{tab:performance_robot}
\resizebox{0.98\textwidth}{!}{
\begin{tabular}{cccccccc}
\noalign{\hrule height 1.2 pt}
\multirow{2}{*}{ Objects } & \multicolumn{2}{c}{ Time efficiency } & & \multicolumn{3}{c}{ Re-identification } & Sweep \\
\cline { 2 - 3 } \cline { 5 - 7 } & Total $(\mathrm{s})$ & Compounding $(\mathrm{s})$ & & Occurrences & Re-identification steps & Success rate & distance (mm) \\
\hline Volunteer 1 & $39(12)$ & $0.41(0.17)$ & & $2.5(1.1)$ & $1.1(0.3)$ & $86 \%$ & $103(14)$ \\
Volunteer 2 & $49(15)$ & $0.54(0.17)$ & & $2.8(1.3)$ & $1.4(0.9)$ & $92 \%$ & $113(9)$ \\
\noalign{\hrule height 1.2 pt}
\end{tabular}
}
\end{table*}

\par
It can be seen from Fig.~\ref{fig:vessel_evaluation} that the re-identification process (case 2) can result in a relatively complete 3D compounding result, while the scan (case 1) without the re-identification process performed poorly. By monitoring the real-time segmentation masks and Doppler images, the system with the re-identification process can be aware of sub-optimal segmentation results. Thereby, automatic probe optimization is performed to achieve good segmentation results for an accurate 3D reconstruction. This characteristic is important to improve the robustness of the automatic US scanning program. Regarding case 2, three re-identification procedures were performed. Qualitatively, the colored geometric distance [see~Fig.~\ref{fig:vessel_evaluation}~(a) and (d)] demonstrate that the distance offsets of some segmentation results in case 1 are larger than $2~mm$, while the maximum offset in Case 2 is around $0.5~mm$.

\par
To further quantitatively compare the results, the Dice Score and IoU are used as performance indicators. The scores for individual frames in the sweeps are depicted in Fig.~\ref{fig:vessel_evaluation}~(b) and (e). The changing tendency in both indicators is consistent with the corresponding distance offset heatmaps. To statistically summarize the performance scores, we group the scores in ten bins ($[0,0.1)$, $[0.1, 0.2)$ ,..., $[0.9,1]$) as can be seen in Fig.~\ref{fig:vessel_evaluation}~(c) and (d). Based on the histogram, there are $204$ frames ($62$ for [0.8,0.9) and $142$ for [0.9, 1.0]) of $242$ frames in total ($84.3\%$) which achieved a high dice score in case 2, while at the same time there are only $88$ frames ($38$ for [0.8,0.9) and $50$ for [0.9, 1.0]) of $183$ respective total frames ($48.1\%$) that achieved a high dice score in case 2. As for the IoU indicator, it has a similar tendency as the dice score. In addition, many results are distributed in the range of $[0,0.5]$ for case 1, while the segmentation scores for case 2 are mainly distributed in the upper part. Finally, the average results are summarized in TABLE~\ref{tab:seg_eval_metrics}. The experimental results demonstrated that the re-identification procedure can effectively improve the robustness and accuracy of the segmentation result (dice score: from $0.54$ to $0.86$; IoU: from $0.47$ to $0.78$).

\subsection{Performance of the Proposed Quality-Aware Blood Vessel Screening Framework}
\par

To further quantitatively analyze the overall performance of the whole framework, we test the proposed quality-aware robotic screening approach on the forearm of two healthy volunteers. Ten independent experiments are carried out on each volunteer (in total $20$ times). Considering the clinical scenarios, we investigate the effectiveness of the Doppler-based re-identification procedure and the time efficiency of the whole scanning. The results have been summarized in TABLE~\ref{tab:performance_robot}.

In order to evaluate the efficacy of the proposed method, experiments were conducted in which the start and end points of the scan trajectory were set manually. The average sweep distance of volunteer 1 and 2 were $103\pm14mm$ and $113\pm9mm$, respectively. The time efficiency was analyzed by measuring the compounding time for the 3D reconstruction and the total time required for the entire workflow. The greater sweep distance for volunteer 2 resulted in a longer scanning time ($49\pm15s$) compared to volunteer 1 ($39\pm12s$), and a proportional increase in compounding time ($0.54\pm0.17s$ vs $0.41\pm0.17s$). Overall, the experiments demonstrated that the robotic scans can be completed within a minute, meeting the standard of clinical practices in terms of time efficiency.

To validate the necessity of the re-identification procedure, we count how often the procedure was activated during scans and also how many out-of-plane adjustment steps were necessary to re-identify a good Doppler signal. It can be seen from TABLE-IV that the average occurrences and required steps in each re-identification procedure on volunteer 1 and volunteer 2 are $2.5\pm1.1$ times vs $2.8\pm1.3$ times and $1.1\pm0.3$ times vs $1.4\pm0.9$ times, respectively. These statistical numbers obtained independently from two different volunteers are very similar. The re-identification procedure was activated over two times on average per scan for both volunteers, emphasizing its crucial role in order to maintain a high segmentation performance.
To ensure the accuracy and completeness of the reconstructed vessel (see Fig.~\ref{fig:vessel_evaluation}), the Doppler-based re-identification is activated. The results demonstrate that after less than two steps of out-of-plane adjustment ($1.1$ and $1.4$ for volunteers 1 and 2, respectively), the system can recover a good Doppler signal. In addition, if the system cannot revive good Doppler results after five searching steps, we consider the re-identification procedure failed. The success rate is $86\%$ (successful / total  = 22/25) and $92\%$ (26/28) on volunteers 1 and 2, respectively. Factors affecting the success rate can include variations in the positioning of the probe over the artery, differences in blood pressure which affects the blood speed, and variations in tissue depth and composition above the arteries. The experimental results demonstrate that the presented re-identification procedure can effectively and quickly adjust the probe to relocate to the orientation leading to good Doppler signal quality.


\section{Discussion}
\par
We present a novel quality-aware robotic US screening framework for tubular structures. Leveraging the Doppler signal, the vascular segmentation performance is improved using the proposed DopUS-Net, and the accuracy and completeness of the reconstructed 3D vessel of interest are significantly enhanced on multiple volunteers. The novel structure of DopUS-Net can also inspire other studies with multiple inputs, such as combining the frequency and image domain with OCT image segmentation~\cite{farshad2022net}. However, there are some limitations that are worth noting. First, the patient's motion is not considered in this work, but such motion could result in disconnected blood vessels in the 3D view. To further address this challenge, previous work on motion-aware RUSS~\cite{jiang2021motion, jiang2022precise, jiang2022towards} could be integrated to monitor and compensate for comparatively large patient motion. Then advanced computational 3D compounding methods~\cite{hennersperger2015computational, hung2021good} need to be developed to tackle the pixel-wise misplacement to guarantee local continuities.
Second, the re-identification is only performed in the out-of-plane direction to recover the Doppler signal in this work. This is the compromised result to reduce the complexity of the re-identification solution, thereby ensuring the time efficiency to meet the requirement for further clinical translation. In the future, if we can access the low-level control of the used US machine, another promising solution is to maintain the probe normal to the constraint surface and directly change the emitted US wave direction. This will further improve the time efficiency by eliminating the need to adjust the probe orientation physically while maintaining stable contact between the probe and the surface. However, it must be noted that this approach requires specialized transducers and cannot be achieved using standard US probes.

\par
In addition, data collection and labeling are often burdensome and expensive tasks, particularly in the field of medical image processing. To improve data efficiency, data augmentation is one of the most popular methods used~\cite{zhang2020generalizing}. To improve the generalization capability of a trained model on unseen data, representation disentanglement~\cite{bi2023mi, meng2020mutual} is investigated to explicitly disentangle the feature in latent space against the domain shift. This is particularly useful for US images because US images are very sensitive to the machine, US machine setting, and also real-time contact conditions. In addition, considering the shortage of data, an emerging concept of federated learning is proposed to enable the training across multiple decentralized servers holding local data examples~\cite{bercea2022federated}. 
Currently, real-time segmentation of ultrasound images is mainly performed on 2D images due to their superior accuracy in capturing local features like boundaries. However, once 3D ultrasound probes generate high-quality volumetric images, 3D segmentation algorithms~\cite{wang2021spherical, thomas2019kpconv} can potentially improve segmentation performance by maintaining the accuracy of local features and enabling anatomical continuity in three dimensions.
\section{Conclusion}
\par
This work presents a novel approach using duplex images to facilitate the accurate segmentation of small blood vessels from cross-sectional US images on volunteers' limbs. To explore the most effective way to incorporate the Doppler signals, various versions of the DopUS-Net have been proposed and compared to the  U-Net~\cite{ronneberger2015u} and VesNet~\cite{jiang2021automatic_baichuan} in terms of dice score. The final version of DopUS-Net has two encoders with different inputs (two-channel ``BD" and pure ``D" images), and two convGRUs applied at the bottleneck layer to exploit the continuity properties of the anatomy from previous frames. In addition, based on the Doppler signal, an online artery re-identification module is developed to qualitatively assess the performance of the real-time image segmentation. Thereby, the quality-aware robotic screening program is developed to further improve the confidence and robustness of the image segmentation. The experimental results demonstrate that the overall performance of the scan with the re-identification process is significantly improved over the scan without the re-identification process on the same volunteer (dice score: from $0.54$ to $0.86$; IoU: from $0.47$ to $0.78$). The Doppler-based quality-aware screening significantly improved the accuracy and robustness of the 3D compounding results; thereby, we \final{anticipate that} it may further improve the acceptance of the autonomous US scanning programs in the future. 

\section*{ACKNOWLEDGMENT}
The authors would like to acknowledge Dr. Med. Reza Ghotbi and Dr. Med. Angelos Karlas for their insightful discussions and valuable clinical feedback. Besides, we want to thank the Editors and reviewers for their implicit contributions to the improvement of this article.

\bibliographystyle{IEEEtran}
\bibliography{IEEEabrv,references}

\vspace{-1cm}
\begin{IEEEbiography}[{\includegraphics[width=1in,height=1.25in,clip,keepaspectratio]{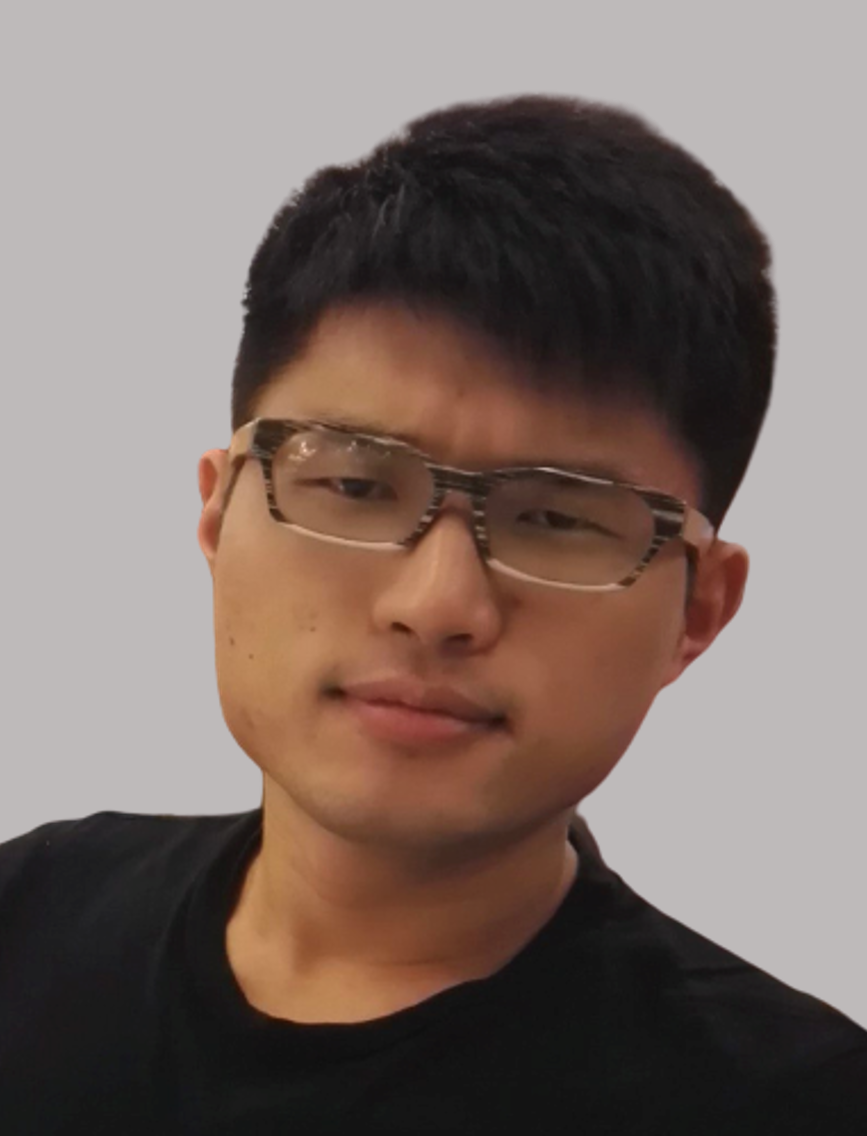}}]
{Zhongliang Jiang} (Member, IEEE) received the M.Eng. degree in Mechanical Engineering from the Harbin Institute of Technology, Shenzhen, China, in 2017, and Ph.D. degree in computer science from the Technical University of Munich, Munich, Germany, in 2022. From January 2017 to July 2018, he worked as a research assistant in the Shenzhen Institutes of Advanced Technology (SIAT) of the Chinese Academy of Science (CAS), Shenzhen, China. He is currently a senior research scientist at the Chair for Computer Aided
Medical Procedures (CAMP) at the Technical University of Munich.

His research interests include medical robotics, robotic learning, human-robot interaction, and robotic ultrasound.
\\
\end{IEEEbiography}

\vspace{-1cm}
\begin{IEEEbiography}
[{\includegraphics[width=1in,height=1.25in,clip,keepaspectratio]{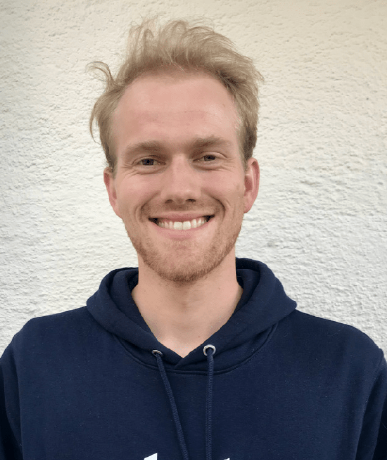}}]
{Felix Duelmer} received the B.Sc. degree in Mechanical Engineering in 2018, and M.Sc. degree in Mechatronics and Robotics in 2023, from the Technical University of Munich, Munich, Germany. He is currently working toward the Ph.D. degree in Computer Science with the Technical University of Munich, Munich, Germany.

His research interests include robotic ultrasound, imaging process, robotic learning, and computer vision.
\\ 
\end{IEEEbiography}

\vspace{-1cm}
\begin{IEEEbiography}[{\includegraphics[width=1in,height=1.25in,clip,keepaspectratio]{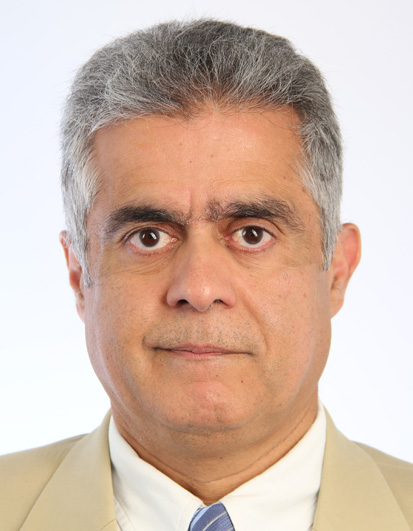}}]
{Nassir Navab} (Fellow, IEEE) received the Ph.D. degree in computer and automation with INRIA, and the University of Paris XI, Paris, France, in 1993.

He is currently a Full Professor and the Director of the Laboratory for Computer-Aided Medical Procedures with the Technical University of Munich, Munich, Germany, and an adjunct professor at Johns Hopkins University, Baltimore, MD, USA. He has also secondary faculty appointments with the both affiliated Medical Schools. He enjoyed two years of a Postdoctoral Fellowship with the MIT Media Laboratory, Cambridge, MA, USA, before joining Siemens Corporate Research (SCR), Princeton, NJ, USA, in 1994. 

Dr. Navab is a fellow of the Academy of Europe, MICCAI, IEEE, and Asia-Pacific Artificial Intelligence Association (AAIA). He was a Distinguished Member and was the recipient of the Siemens Inventor of the Year Award in 2001, at SCR, the SMIT Society Technology award in 2010 for the introduction of Camera Augmented Mobile C-arm and Freehand SPECT technologies, and the ``$10$ years lasting impact award" of IEEE ISMAR in 2015. He is the author of hundreds of peer-reviewed scientific papers, with more than 54,400 citations and enjoy an h-index of 104 as of August 11, 2022. He is the author of more than thirty awarded papers including 11 at MICCAI, 5 at IPCAI, and three at IEEE ISMAR. He is the inventor of 50 granted US patents and more than 50 International ones.
\\ \\ 
\end{IEEEbiography}

\end{document}